\pdfoutput=1

\documentclass[11pt]{article}

\usepackage{COLING2025}

\usepackage{times}
\usepackage{latexsym}
\usepackage{amsmath}
\usepackage{graphicx}
\usepackage{booktabs}
\usepackage{tabularray}
\usepackage{array}
\usepackage{ragged2e}
\usepackage{multirow}
\usepackage{amsfonts}
\usepackage{makecell}
\usepackage{colortbl}      
\usepackage[ruled,vlined]{algorithm2e}
\usepackage{setspace}
\usepackage{hyperref}
\usepackage{caption} 
\usepackage{subcaption} 
\usepackage[english]{babel}
\usepackage{wasysym}
\usepackage{longtable}
\usepackage{array}
\usepackage{pdfpages}
\usepackage{float}
\usepackage{hyperref}

\definecolor{commentcolor}{RGB}{110,154,155}   

\usepackage[T1]{fontenc}

\usepackage[utf8]{inputenc}

\usepackage{microtype}

\title{Fine-tuning Large Language Models for Improving Factuality in Legal Question Answering}

\author{
  Yinghao Hu\textsuperscript{1},
  Leilei Gan\textsuperscript{1, *},
  Wenyi Xiao\textsuperscript{1},
  Kun Kuang\textsuperscript{2},
  Fei Wu\textsuperscript{2} \\
  \texttt{\{huyinghao, leileigan, wenyixiao, kunkuang, wufei\}@zju.edu.cn}, \\
  \textsuperscript{1}School of Software Technology, Zhejiang University \\
  \textsuperscript{2}College of Computer Science and Technology, Zhejiang University \\
}

\begin{document}
\maketitle

\renewcommand\thefootnote{*}
\footnotetext{Corresponding author}
\renewcommand\thefootnote{\arabic{footnote}}

\begin{abstract}

Hallucination, or the generation of incorrect or fabricated information, remains a critical challenge in large language models (LLMs), particularly in high-stake domains such as legal question answering (QA). In order to mitigate the hallucination rate in legal QA, we first introduce a benchmark called LegalHalBench and three automatic metrics to evaluate the common hallucinations when LLMs answer legal questions. We then propose a hallucination mitigation method that integrates behavior cloning and a novel Hard Sample-aware Iterative Direct Preference Optimization (HIPO). We conduct extensive real-data experiments to validate the effectiveness of our approach. Our results demonstrate remarkable improvements in various metrics, including the newly proposed Non-Hallucinated Statute Rate, Statute Relevance Rate, Legal Claim Truthfulness, as well as traditional metrics such as METEOR, BERTScore, ROUGE-L, and win rates. Our benchmark and model are available at \url{https://github.com/YinghaoHu/LegalHalBench}.

\end{abstract}

\section{Introduction}
\begin{figure}[t]
    \centering
    \includegraphics[width=\columnwidth, trim=5 5 5 5, clip]{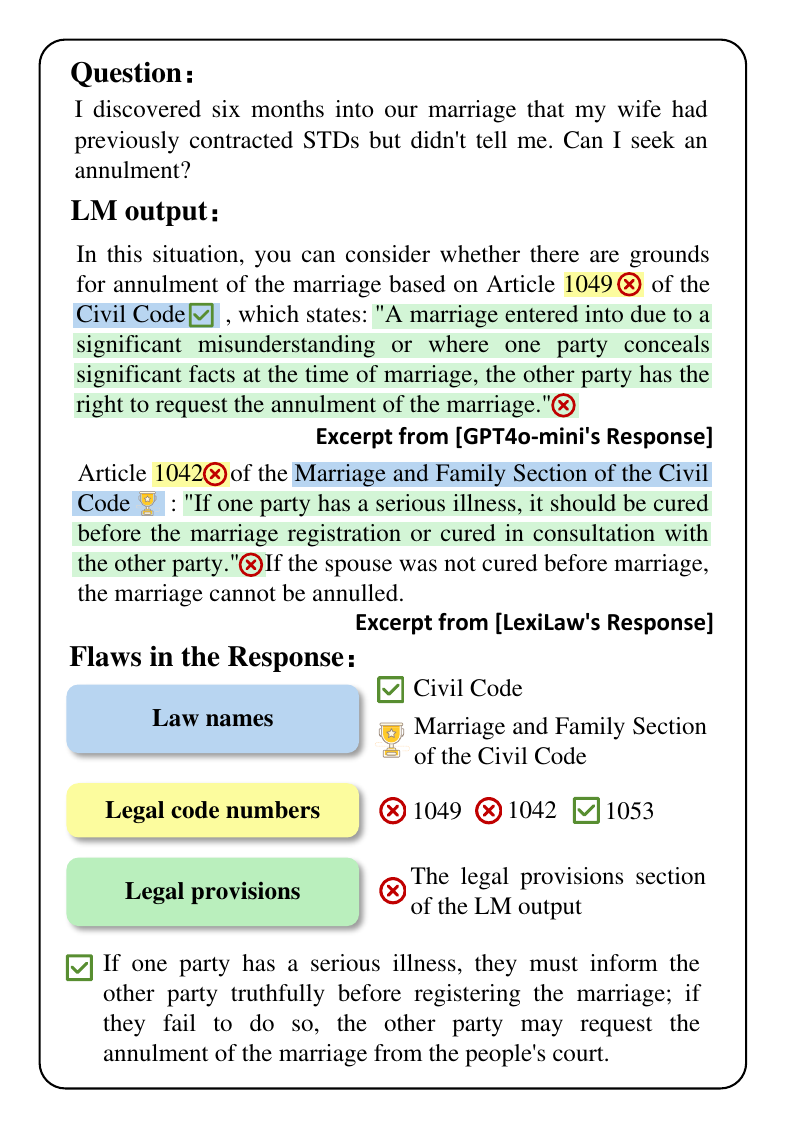}
    \caption{Hallucinations of LLMs in the legal question answering task.}
    \label{fig:motivation}
\end{figure}

Recently, large language models (LLMs) have demonstrated notable advancement and have been applied to a broad spectrum of natural language process (NLP) tasks across different domains, such as medical~\cite{zhou2023medicalsurvey}, financial~\cite{wang2023fingpt} and legal domains~\cite{trautmann2022legal}. Among them, the application of LLMs in the legal field can help legal practitioners substantially improve their productivity in completing daunting tasks such as legal judgment prediction~\cite{gan2021judgment, masala2024improving, li2024graph}, legal question answering~\cite{gan2021dialogue, louis2024interpretable}, generation of judicial opinions~\cite{zhou2024unlocking, li2024enhancing, liu2024unleashing} and legal document drafting~\cite{blair2023can,choi2021chatgpt,hargreaves2023words}. 

Despite showing promising benefits, general LLMs are usually troubled by the so-called hallucination problem which refers to the scenario of LLM generating responses that are inconsistent with the real-world legal facts~\cite{dahl2024large}. To improve legal factuality of LLMs, some specialized LLMs for the legal domain have been proposed~\cite{pengxiaosong2023lawgpt,haitaoli2023LexiLaw,wisdomInterrogatory} by either continually training general LLMs~\cite{wisdomInterrogatory} on massive legal datasets or by employing retrieval-augmented generation technique~\cite{huang2023lawyer}. In this study, we focus on the former approach as it can inherently mitigate hallucinations by injecting legal knowledge into the models. Successful progress has been made in this direction. For example, the Lawyer LLaMA~\cite{huang2023lawyer} model introduces a legal statute retriever for marriage-related issues, while ChatLaw2-MoE~\cite{ChatLaw} employs a mixture of experts (MoE) model and a multi-agent system to improve factuality for legal domain applications. 

However, the issue of hallucination in LLMs within the legal domain, particularly in the task of legal question answering (QA), remains a significant challenge. Legal QA requires not only the memorization of legal knowledge but also the ability to synthesize this knowledge into factual responses. For example, as shown in Fig.~\ref{fig:motivation}, when answering a legal question, both the general LLM GPT-4o-mini and the specialized LLM LexiLaw models make fabricated content on the citation of law provisions and generate claims which were not in line with the actual law provisions. The inaccuracy of these LLM-based legal QA substantially impedes their practical application, as the high-stakes legal domain demands faithful and accurate interpretations of the law, along with harm-free and precise legal advice. Surprisingly, few studies have thoroughly addressed the LLM hallucination problem in legal QA. Moreover, there is a clear lack of tailored hallucination evaluation metrics and benchmarks in the literature.

In this paper, to improve the factuality in legal QA, we first develop a legal hallucination evaluation benchmark (LegalHalBench), including various automatic metrics for detecting five types of common hallucinations which are often generated by LLMs. Then, a hallucination mitigation method is proposed to help LLMs cite correct law provisions and generate factual claims. This method includes two stages: (1) a supervised fine-tuning (SFT) stage and (2) a hard sample-aware iterative direct preference optimization stage. Current studies have demonstrated that both the SFT~\cite{huang2023survey} and preference learning techniques ~\cite{saeidi2024insights,schulman2023reinforcement} can help mitigate hallucinations in LLMs. Additionally, in order to generate a large-scale legal QA dataset for training LLMs, we propose an automatic legal QA dataset curation approach to help alleviate the manual annotation cost and improve the scalablity of the datasets. 

We conduct extensive experiments on the curated LegalHalBench using our newly introduced hallucination evaluation metrics. The experimental results demonstrate the effectiveness of the proposed two-stage fine-tuning strategy. Our method achieves a non-hallucinated statute rate of 38.353\%, significantly surpassing specialized LLMs such as WisdomInterrogatory and LexiLaw, as well as general models like Llama3.1 405B and GPT-4o. Furthermore, our model exhibits improvements of 37.13\% in statute relevance rate and 6.56\% in legal claim truthfulness compared to the vanilla version of the base model. Additionally, helpfulness evaluation experiments reveal that our method achieves a dominant win rate compared to existing legal LLMs and general-purpose LLMs. The meticulously designed ablation studies further demonstrate the effectiveness of each proposed component. The human consistency experiment, conducted to evaluate the legal hallucination metrics, supports the reliability of our proposed metrics.

\section{Related Work}
\subsection{Large Language Models in Legal Domain}
Based on the idea of LLM training, various pre-trained LLMs for legal QA have been proposed to help general LLMs comprehend legal terminology and learn legal knowledge~\cite{chalkidis2020legal,zheng2021does,pengxiaosong2023lawgpt,haitaoli2023LexiLaw,huang2023lawyer,ChatLaw}. After training, these specialized models have been employed to complete various legal tasks such as predicting legal judgments, analyzing legal documents, and drafting legal texts~\cite{iu2023chatgpt,macey2023chatgpt,oltz2023chatgpt,gan2023exploiting}. For example, LawGPT~\cite{pengxiaosong2023lawgpt} is built by continually training Llama~\cite{touvron2023llama1} on extensive legal QA datasets. Based on ChatGLM-6B, LexiLaw~\cite{haitaoli2023LexiLaw} is trained on a mixup dataset from general domain and legal domain.  Nevertheless, these legal-specific LLMs such as LawGPT, LexiLaw, Lawyer LLaMA, and ChatLaw, as well as the general-purpose LLMs like Qwen2~\cite{yang2024qwen2}, GLM4~\cite{glm2024chatglm}, GPT-4~\cite{openai2024gpt4}, and Llama 3.1~\cite{dubey2024llama3herdmodels} still occasionally produce misleading or erroneous responses.

\subsection{Hallucinations in Large Language Models}
LLM Hallucination~\cite{schulman2023reinforcement, zhang2023hallucination} refers to the phenomenon where the generated responses of LLMs are factually incorrect, or not grounded with given contexts. To alleviate the LLM hallucination problem, there are mainly two lines of research work.  One is based on the retrieval-augmented generation techniques~\cite{lewis2020retrieval,guu2020retrieval} which can retrieve relevant information to address the knowledge gap. The other is training-based approaches which seek to reduce hallucinations in LLMs via further training, such as instruction fine-tuning~\cite{elaraby2023halo} or preference learning~\cite{tian2023fine,yuan2023rrhfrankresponsesalign,ethayarajh2024ktomodelalignmentprospect,xiao2024comprehensive, xiao2024detecting}. For example, ~\citet{elaraby2023halo} proposes to use an SFT model with a curated, domain-specific dataset to alleviate hallucinations. In preference learning, Proximal Policy Optimization (PPO)~\cite{schulman2017proximal} and Direct Preference Optimization (DPO)~\cite{rafailov2023direct} have been exploited to enhance the factuality of LLMs~\cite{lee2022factuality} as well as various adaptations, including Self-Rewarding~\cite{yuan2024self}, Iterative DPO~\cite{xu2023some,xiong2023gibbs}, and SimPO~\cite{meng2024simpo}. In this work, we particularly focus on the factuality issues of the LLMs in legal QA and propose a two-stage fine-tuning model combining the SFT and hard sample-aware iterative DPO techniques to effectively mitigate hallucinations of the LLMs.

\section{Legal Hallucination Definition, Benchmark and Evaluation Metrics}
To improve the factuality of the LLMs when addressing legal queries, we first introduce five commonly generated hallucination types. Then, we design the respective metrics for evaluating these specific hallucinations.

\subsection{Hallucinations in Legal Question Answering}
\label{sec:DefineHallucination}

We define five hallucination types which are often generated by LLMs.
\paragraph{Incorrect law name.}This error occurs when the name of a law is incorrectly stated. For example, mistakenly referring to the "Clean Air Act" as the "Air Protection Act."
\paragraph{Incorrect legal code number.}This happens when a specific section or provision of the law is cited with the wrong number. For example, citing "Section 302" when it should be "Section 303."
\paragraph{Fabrication of legal provision.}This involves making up a law or legal provision that does not actually exist. For example, referencing a non-existent "Article 15 on Data Privacy" in a law where no such article exists.
\paragraph{Incorrect citation of legal provision.}This type of error occurs when the cited legal provision is correct but irrelevant to the issue at hand. For example, citing a provision about road safety in a question that pertains to inheritance.
\paragraph{Suggestions that contradict regulations.}This type of error involves providing advice or recommendations that directly contradict the existing legal rules or regulations, for example, proposing a course of action that is explicitly prohibited by the law.

\subsection{Legal Hallucination Benchmark}
\label{sec:LegalHallucinationBenchmark}
To thoroughly evaluate the phenomenon of hallucination in LLMs when addressing legal queries, we introduce the first benchmark for legal hallucination detection, designated LegalHalBench. This benchmark encompasses a range of scenarios from both civil and criminal law, including \textit{Inheritance Law}, \textit{Road Traffic Safety Law}, \textit{Marriage Law}, \textit{Tort Liability Law}, \textit{Real Rights Law}, \textit{Lending Law}, and \textit{Criminal Law}. These laws are selected based on their prevalence according to the statistical distribution reported by~\citet{chen2023equals}.

The questions and reference answers in LegalHalBench are constructed as follows: (1) For civil questions, we primarily gather inquiries from phone consultations and online queries conducted at a law firm. The reference answers are initially generated by an LLM (e.g., GPT-4-turbo) and subsequently subjected to rigorous review and refinement by legal experts. Next, legal professionals are asked to provide authoritative legal statutes to be included in the reference answers. Additionally, questions shorter than 20 words are filtered out, and unclear inquiries are rephrased to ensure they are answerable. (2) For criminal questions, we source them from criminal judgment documents issued by the courts. Given the highly structured nature of criminal judgments, we can directly use regular expressions to extract necessary question information from sections such as "Defendant’s Basic Information" and "Facts of the Prosecution", and answers from sections like "Court’s Opinion" and "Judgment", including relevant legal articles and judgment outcomes, respectively.  The extracted information is then combined with our designed instructions. 

Finally, LegalHalBench consists of a total of 1,988 questions, covering over 800 distinct legal provisions. The dataset is structured in the format {question, reference statutes, reference answers}. The statistical distribution of LegalHalBench is presented in Tab.~\ref{DataDistributionofLegalHalBench}. The cost of constructing this dataset is detailed in Sec.~\ref{sec:CostofConstructingtheLegalHalBench}.

\subsection{Legal Hallucination Evaluation Metrics}
\label{sec:metrics}
Manually evaluating model-generated responses for hallucinations is costly and time-consuming. Inspired by~\citet{min2023factscore}, we propose the following metrics to automatically detect hallucinations in the responses. 

\paragraph{Non-Hallucinated Statute Rate.} 
We define the Non-Hallucinated Statute Rate (NHSR) as follows: Let \(\hat{y}\) denote the statutes generated by the model. The NHSR is the proportion of non-hallucinated statutes among all statutes in \(\hat{y}\). A statute is considered non-hallucinated if its name, number, and content are entirely accurate when compared to the golden reference statute.

\begin{equation}
    \text{NHSR} = \frac{\sum_{i=1}^N \mathbb{I}(S_{\text{name},i} \wedge S_{\text{number},i} \wedge S_{\text{content},i})}{N}
\end{equation}
Here, \(\mathbb{I}()\) and $N$ represent an indicator function and the total number of statutes in $\hat{y}$. \(S_{\text{name},i}\), \(S_{\text{number},i}\), and \(S_{\text{content},i}\) represent whether the statute's name, number, and content are correct, respectively. The values of $S_{\text{name},i}$, $S_{\text{number},i}$ and $S_{\text{content},i}$ are obtained by first extracting statutes from $\hat{y}$ via prompting LLMs and then comparing them with each part of the most similar golden statute. A more formal calculation process is provided in the Sec.~\ref{sec:TechnicalDetailsofNHSR}.

\paragraph{Statute Relevance Rate.}
This metric measures the relevance between the knowledge contained in legal statutes and the legal question. We denote this metric as \(\textit{Rel}\). The value of \(\textit{Rel}\) ranges from 0 to 10, where values closer to 0 indicate that the knowledge in the statutes is less relevant to the question, while values closer to 10 suggest a higher relevance. The technical details and calculation formula can be found in the Sec.~\ref{sec:TechnicalDetailsofStatuteRelevanceRate}.

\paragraph{Legal Claim Truthfulness.}
This metric measures the truthfulness of claims in the model-generated answers. We denote this metric as $T_{\text{LC}}$. The \(T_{\text{LC}}\) ranges from 0 to 10, where a score closer to 10 indicates higher truthfulness, and a lower score suggests a greater likelihood of unfounded or incorrect legal claims. The technical details and calculation formula can be found in the Sec.~\ref{sec:TechnicalDetailsofLegalClaimTruthfulness}.

\section{Method}
To improve the factuality of LLMs for legal question answering, we propose a two-stage fine-tuning algorithm, including SFT and hard sample-aware iterative direct preference optimization techniques. Current studies have demonstrated that both SFT~\cite{huang2023survey} and preference learning techniques~\cite{saeidi2024insights,schulman2023reinforcement} can help mitigate LLM hallucinations.

\subsection{Training Dataset Construction}

\begin{figure*}[htbp]
    \centering
    \includegraphics[width=\textwidth, trim=5 15 5 15, clip]{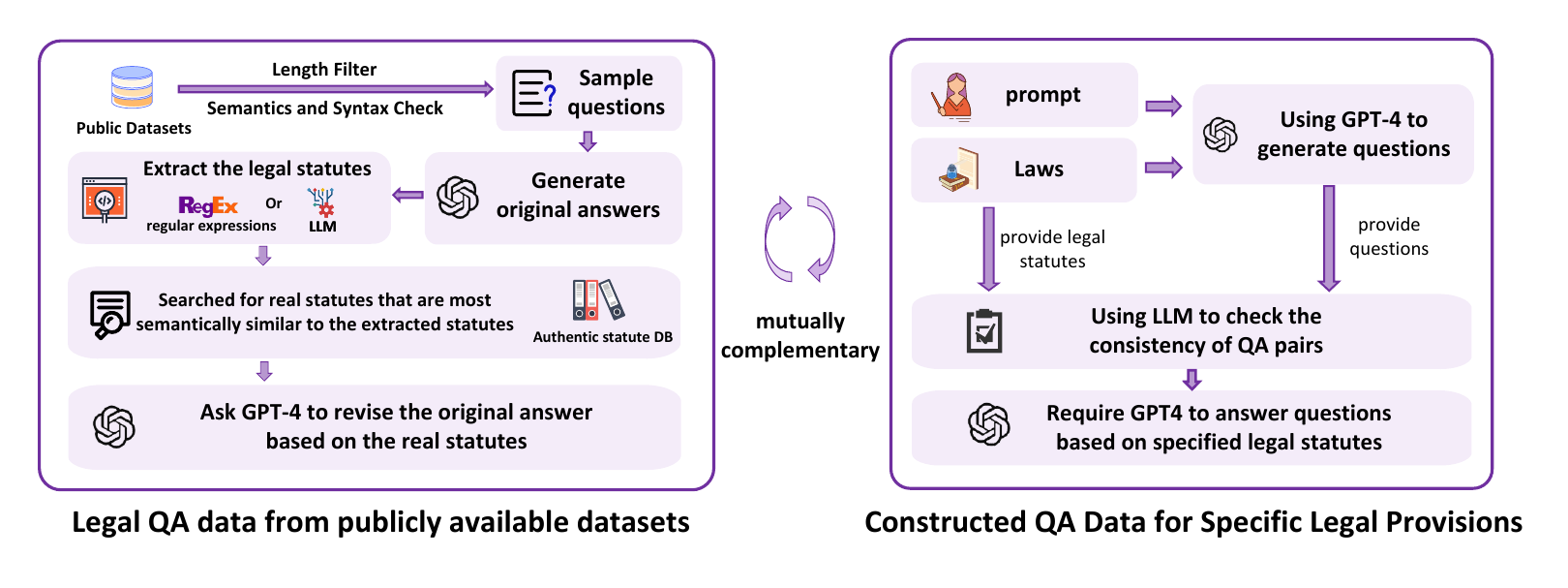}
    \caption{An illustration of the training dataset construction pipeline.}
    \label{fig:ConstructingData}
\end{figure*}

At present, there is a lack of large-scale QA datasets that include accurate citations of legal provisions suitable for fine-tuning. Therefore, as illustrated in Fig.~\ref{fig:ConstructingData}, we propose a two-step automated construction method to initially curate such a dataset, which is not only cost-effective but also scalable.

\paragraph{Legal QA data from publicly available datasets.}
\label{sec:QAforpubliclydatasets} 
We use the CAIL2018 legal QA dataset~\cite{xiao2018cail2018} as one of the data source, which comprises questions derived from real-world scenarios. However, we find that the provided answers are neither informative nor cite relevant statutes. To address this issue, we turn to proprietary LLMs to generate high-quality responses, as shown in Fig.~\ref{fig:ConstructingData}. Nonetheless, due to lack of legal knowledge, proprietary LLMs such as GPT-4-turbo may also fabricate legal statutes. To mitigate this, we first extract statutes from the LLM-generated response and use these extracted statutes as queries to search the most similar statute in an authentic statute database. We then prompt GPT-4-turbo with the most similar statute to revise its generated response accordingly. Detailed prompts for extracting statute are provided in Sec.~\ref{sec:ExtractingStatutesUsingLLMs}.

Through the aforementioned procedures, we obtain 12,149 legal QA samples with answers that are not only informative but also include accurate citations of legal provisions. For a detailed evaluation of the quality of the training data, refer to Sec.~\ref{sec:ExplorationofTrainingDataQuality}.

\paragraph{Legal Provisions based Legal QA dataset Construction.} 
\label{sec:QuestionConstruction}
While the initial portion of the dataset is of high quality, it inherently lacks comprehensive coverage of all legal provisions due to the extensive number of provisions. Additionally, we aim to instruct the model on citing specific legal provisions. To address this, we propose a legal provision-based approach for constructing training data points with the aid of GPT-4-turbo.

Specifically, as shown in Fig.~\ref{fig:ConstructingData}, given a legal provision $L$, we design question generation instruction to prompt GPT-4-turbo to generate question $Q$ which is relevant to the legal provision $L$. Subsequently, using the newly generated question $Q$, the legal provision $L$, we further prompt GPT-4-turbo to generate an answer $A$, which not only adheres to the given question but also avoids hallucinations in the response due to the correct legal provision provided in the context. This approach enables us to freely expand the amount of training data for legal provisions not covered in the first portion. 

Through this process, we collected a total of 3,883 legal QA data points.

\subsection{Training Stage I: Supervised Fine-tuning}
\label{sec:TrainingStageI}
Given the above automatically constructed dataset, in the first stage, we perform supervised fine-tuning to instruct the model to correctly cite law provisions~\cite{gao2023enabling}. The training objective is to minimize the negative log-likelihood (NLL) loss:
\begin{equation}
    \mathcal{L}_{\text{SFT}} = - \sum_{t=1}^{T} \log P(y_t \mid X, y_{<t}; \theta)
\end{equation}
where $T$ is the length of the input and target sequence. $y_t$ is the target token at timestep $t$. $y_{<t}$ represents all tokens before timestep $t$, i.e., $y_1, y_2, \ldots, y_{t-1}$. $P(y_t \mid X, y_{<t}; \theta)$ denotes the probability of the target token $y_t$ given the input sequence $X$ and all previous target tokens $y_{<t}$, modeled by parameters $\theta$.

\subsection{Stage II: Hard Sample-Aware Iterative Direct Preference Optimization}

In stage II, we continue to fine-tune the LLM using preference learning, which has been demonstrated to be effective in mitigating hallucinations\cite{saeidi2024insights,schulman2023reinforcement,tian2023fine}. Inspired by recent advancement in iterative offline RLHF for instruction following~\cite{yuan2024self,xu2023some,pang2024iterative}, we propose a \textbf{H}ard sample-aware \textbf{I}terative direct \textbf{P}reference \textbf{O}ptimization (HIPO) to further mitigate hallucinations, thereby improving the factuality of LLM for legal QA.

\begin{algorithm}[h]
\caption{Hard Sample-aware Iterative Direct Preference Optimization}
\SetAlgoLined
\KwIn{$M_0$: Initial model, $\mathcal{D}_{\text{HIPO}} = \{<x, y_w, y_l>\}$, where \(x\) represents a legal domain question, \(y_w\) is the answer to be learned, sourced from \(\mathcal{D}_{\text{SFT}}\), and \(y_l\) is the rejected answer generated by model \(M_0\).}
\KwResult{$M_t$: Model after $t$ iterations.}

\SetKwProg{Proc}{Procedure}{}{}
\Proc{Iterative Training}{
  \For{$i = 1$ \textbf{to} $t$}{
    \(y_l^i \leftarrow\) GenerateAns($M_{i-1}, x^i$) \\
    \(signal \leftarrow\) CheckHallucination($x^i, y_l^i,\text{metrics}$) \\
    \If{\text{no hallucinations in } $signal$}{
    \(\mathcal{D}_{\text{HIPO}}^{i} \leftarrow\) ConstructTrainingSet($\{x^i, y_w^i, y_l^i\}$) \\
    $M_{i+1} \leftarrow$ TrainDPO($M_i, \mathcal{D}_{\text{HIPO}}^{i}$)
  }
 }
 \KwRet{$M_t$}
}
\end{algorithm}
Given the unique nature of legal tasks, users often prefer that legal provisions be recited verbatim by the LLM. However, a model trained solely on binary preference learning may not effectively accomplish this task~\cite{ethayarajh2024ktomodelalignmentprospect}. Therefore, at the heart of HIPO, we dedicatedly design two features to effectively exploit both the positive and negative signals. First, in our iterative offline prefence learning, we use the positive and negative signals to select training samples for the next round. In other words, after each iteration, we employ the metrics introduced in Sec.~\ref{sec:metrics} to identify whether the generated answers contain hallucinations. Training samples that do not exhibit hallucinations are excluded from the training set. As iterations progress, simpler training samples are continuously filtered out, leaving increasingly challenging samples. 

Second, since the positive examples of the training data largely consists of information that the model has not yet mastered, our objective is to ensure that the model learns no-hallucination statutes and provide helpful answers from the selected responses. Inspired by ~\citet{yuan2023rrhfrankresponsesalign}, ~\citet{xu2024contrastivepreferenceoptimizationpushing} and ~\citet{hong2024orpo}, we use the NLL loss to enhance the model's learning from positive signals provided by these selected responses.

Specifically, at each iteration of HIPO, we begin by preparing training data for that round. We define the dataset for the \(i\)-th iteration as \(\mathcal{D}_{\text{HIPO}}=\{<x^{i}, y_{w}^{i}, y_{l}^{i}>\}\). Here, \(x\) represents a legal domain question, and \(y_{w}\) is the chosen answer sourced from \(\mathcal{D}_{\text{SFT}}\). \(y_{l}\) denotes the rejected answer, which is generated by the model \(M_{i-1}\). We use NHSR in Sec.~\ref{sec:metrics} and BERTScore~\cite{zhang2019bertscore} as the selection metrics \(Q\). An answer will not be included into $\mathcal{D}_{\text{HIPO}}$ if it satisfies: (i) the generated statute has no hallucinations and (ii) the semantic similarity between the generated answer \(y_{resp}\) and the chosen answer \(y_{w}\) exceeds a pre-defined threshold. Once upon \(\mathcal{D}_{\text{HIPO}}\) is prepared, we can proceed to train the model \(M_{i}\) in the current round. The training objective is to minimize the weighted sum of the NLL loss and the DPO loss, thereby effectively utilizing both positive and negative signals.
\begin{equation}
\begin{aligned}
& \mathcal{L}_{\text{HIPO}}=\mathcal{I}\left(y_l\right) \cdot \mathcal{L}_{\text{NLL}}\left(y_w \mid x\right)+\mathcal{L}_{\text{DPO}}\left(y_w, y_l \mid x\right) \\
& \quad=-\mathcal{I}\left(y_l\right) \cdot \frac{\log M_\theta\left(y_w \mid x\right)}{\left|y_w\right|} \\
& -\log \sigma\left(\beta \frac{\log M_\theta\left(y_w \mid x\right)}{\log M_t\left(y_w \mid x\right)}-\beta \frac{\log M_\theta\left(y_l \mid x\right)}{\log M_t\left(y_l \mid x\right)}\right) 
\end{aligned}
\label{eq:loss}
\end{equation}
Here, $M(x)$ denotes the probability of sequence $x$ under the model $M$, and $\mathcal{I}(y_l) $ is the indicator function of statutes hallucination rate over $y_l$. We use the previous iteration's model $M_t$ as the reference model in the denominator of the DPO term. Note that we omit the reducing iteration index for brevity.

\section{Experiments}

\subsection{Experimental Setup}
\paragraph{Datasets.} We conduct experiments to evaluate the legal QA capabilities of our method and baselines on the LegalHallBench which is introduced in Sec.~\ref{sec:LegalHallucinationBenchmark}. We also conduct tests on some tasks of the Lawbench~\cite{fei2023lawbench} dataset, as detailed in Sec.~\ref{sec:EvaluatingModelPerformanceonLawBench}.

\paragraph{Baselines.}
We compare our model with the following baseline LLMs in legal QA: Open-source general-purpose LLMs, including Qwen2-Instruct-7B~\cite{yang2024qwen2}, GLM4-Chat-9B~\cite{glm2024chatglm}, Llama 3.1-70B, and Llama 3.1-405B~\cite{dubey2024llama}. Closed-source general-purpose LLMs, including GPT-4o-mini and GPT-4o. Legal-specific LLMs, including wisdomInterrogatory~\cite{wisdomInterrogatory} and Lexilaw~\cite{haitaoli2023LexiLaw}. In some models, we also integrate the BGE~\cite{bge_embedding} retriever to enhance the strength of the baseline. More details about the baselines are reported in Sec.~\ref{sec:DetailedIntroductiontoBaseline}.

\paragraph{Metrics.}
For LegalHalBench, we use the metrics introduced in Sec.~\ref{sec:metrics} to evaluate the factuality performance of LLMs for legal QA. Additionally, we select METEOR~\cite{banerjee2005meteor}, BERTScore~\cite{zhang2019bertscore}, and Rouge-L~\cite{lin2004rouge} as evaluation metrics to assess the differences between the model-generated answers and the reference answers. Furthermore, we use the HIPO round 3 version of the model based on the GLM4 foundation as the reference model. We conduct pairwise comparisons between the responses generated by all models and the responses generated by this reference model. GPT-4-turbo serves as the evaluator, calculating the win rate based on its preference for the responses from different models.

\paragraph{Implementation Details.}
Further details about the implementation can be found in Sec.~\ref{sec:ImplementationDetails}.

\subsection{Main Results}
We report the results regarding factuality and overall helpfulness as follows. Notably, using the HIPO training strategy, we conduct three rounds of iteration. Ultimately, we select the model from the third round for subsequent experimental comparisons based on its optimal performance.

\paragraph{Factuality Results.} 

\begin{table}[t]
\small
\renewcommand{\arraystretch}{1}
\begin{tabular}{>{\raggedright\arraybackslash}m{2.2cm}>{\centering\arraybackslash}m{1.3cm}>{\centering\arraybackslash}m{1.1cm}>{\centering\arraybackslash}m{1.1cm}}  
\toprule
\textbf{Models} & \textbf{NHSR} & \textbf{\textit{Rel}} & \textbf{T$_{\text{LC}}$}\\
\hline
\multicolumn{4}{c}{\textit{\underline{Open-source LLMs}}} \\
Llama3.1 70B & 1.595\% & 6.370 & 8.372 \\
Llama3.1 405B & 5.036\% & 6.625 & 8.741 \\
\hline
\multicolumn{4}{c}{\textit{\underline{Proprietary LLMs}}} \\
GPT4o-mini & 0.806\% & 6.422 & 8.875 \\
GPT4o & 16.029\% & 6.953 & \textbf{9.300} \\
\hline
\multicolumn{4}{c}{\textit{\underline{Specialized Legal LLMs}}} \\
wisdomInt. & 18.089\% & 5.444 & 8.496 \\
Lexilaw & 17.618\% & 6.487 & 8.488 \\
\hline
\multicolumn{4}{c}{\textit{\underline{Qwen2 Instruct 7B}}} \\
Vanilla & 7.501\% & 5.633 & 8.571 \\
w/ SFT & 24.239\% & 5.179 & 8.592 \\
w/ SFT + DPO & 24.693\% & 5.764 & 8.830 \\
w/ SFT + SimPO & 28.492\% & 5.527 & 8.720 \\
w/ SFT + HIPO & 27.435\% & 5.929 & \underline{9.181} \\
\hline
\multicolumn{4}{c}{\textit{\underline{GLM4 Chat 9B}}} \\
Vanilla & 6.541\% & 5.123 & 8.520 \\
w/ SFT & 26.941\% & 5.832 & 8.771 \\
w/ SFT + DPO & 28.691\% & 6.895 & 9.042 \\
w/ SFT + SimPO & \underline{31.402\%} & \underline{6.975} & 9.072 \\
w/ SFT + HIPO  & \textbf{38.353\%} & \textbf{7.025} & 9.079 \\
\bottomrule
\end{tabular}
\caption{Factuality results on LegalHalBench. We conduct experiments on different LLMs. Bold scores represent the best performance within the same model, while underlined scores represent the second best.}
\label{FactualityResultsTable}
\end{table}

As shown in Tab.~\ref{FactualityResultsTable}, the models trained using the HIPO strategy demonstrate strong performance in non-hallucinated statute rate on both Qwen2 and GLM4 bases, with improvements of 19.934\% and 31.812\% over the respective base models. We also observe that the legal-specific LLMs achieve higher rates due to their training on legal data, surpassing models like Llama3.1-70B, Llama3.1-405B, GPT4o-mini, and GPT4o. When analyzing the statute relevance rate, it is observed that models tend to exhibit higher useful knowledge in their outputs as the size of their parameters increases. Additionally, HIPO training significantly improves the statute relevance rate, particularly on GLM4, with an increase of 1.902, achieving a 37.13\% improvement over the base model, surpassing all baselines. Regarding legal claim truthfulness, our trained models outperform many legal-specific and general-purpose LLMs, although they slightly lag behind GPT4o, which excels due to its extensive world knowledge and conservative approach in legal contexts.

\paragraph{Helpfulness Results.} 

\begin{table}[t]
\centering
\small
\begin{tabular}{
>{\raggedright\arraybackslash}m{2.2cm}
>{\centering\arraybackslash}m{1.3cm}
>{\centering\arraybackslash}m{1.3cm}
>{\centering\arraybackslash}m{1.4cm}}  
\toprule
\textbf{Models} & \textbf{METEOR} & \textbf{BERTScore}  &\textbf{ROUGE} \\
\hline
\multicolumn{4}{c}{\textit{\underline{Open-source LLMs}}} \\
Llama3.1 70B & 0.200 & 0.707 & 0.221 \\
Llama3.1 405B & 0.232 & 0.726 & 0.249 \\
\hline
\multicolumn{4}{c}{\textit{\underline{Proprietary LLMs}}} \\
GPT4o-mini & 0.290 & 0.731 & 0.254 \\
GPT4o & 0.348 & 0.739 & 0.266 \\
\hline
\multicolumn{4}{c}{\textit{\underline{Specialized Legal LLMs}}} \\
wisdomInt. & 0.303 & 0.721 & 0.261\\
Lexilaw & 0.186 & 0.718 & 0.234 \\
\hline
\multicolumn{4}{c}{\textit{\underline{Qwen2 Instruct 7B}}} \\
Vanilla & 0.275 & 0.692 & 0.176 \\
w/ SFT & 0.317 & 0.712 & 0.228 \\
w/ SFT + DPO & 0.371 & 0.743 & 0.295 \\
w/ SFT + SimPO & 0.342 & 0.725 & 0.258 \\
w/ SFT + HIPO & 0.366 & 0.746 & 0.303 \\
\hline
\multicolumn{4}{c}{\textit{\underline{GLM4 Chat 9B}}} \\
Vanilla & 0.285 & 0.710 & 0.150 \\
w/ SFT & 0.329 & 0.744 & 0.304 \\
w/ SFT + DPO & \underline{0.406} &  \textbf{0.762} & 0.333 \\
w/ SFT + SimPO & 0.396 & 0.760 & \underline{0.338} \\
w/ SFT + HIPO & \textbf{0.407} & \underline{0.762} & \textbf{0.340} \\
\bottomrule
\end{tabular}
\caption{Helpfulness results on LegalHalBench. We conduct experiments on different LLMs. Bold scores represent the best performance within the same model, while underlined scores represent the second best.}
\label{HelpfulnessResultsTable}
\end{table}

As demonstrated in Tab.~\ref{HelpfulnessResultsTable}, without the use of external knowledge, the combination of SFT and HIPO on the GLM4-Chat-9B model significantly enhances performance on three metrics: METEOR, BERTScore, and ROUGE-L, showing improvements of 42.8\%, 7.3\%, and 126.7\% over the baseline, respectively. This performance surpasses all existing models on these metrics.

\subsection{Analyzes}

\paragraph{Impact of Different HIPO Iterations.}

Tab.~\ref{ResultsTable_HIPOIterations} presents the impact of HIPO iterations on the base models Qwen2-Instruct-7B and GLM4-Chat-9B. We find that on both base models, performance improvements across various metrics tend to stabilize after the third HIPO iteration. This observation is consistent with the results reported in ~\citet{pang2024iterative} and ~\citet{yuan2024self}.

\begin{table}[t]
\centering
\small
\begin{tabular}{
>{\raggedright\arraybackslash}m{1.5cm}
>{\centering\arraybackslash}m{0.9cm}
>{\centering\arraybackslash}m{0.6cm}
>{\centering\arraybackslash}m{0.6cm}
>{\centering\arraybackslash}m{1cm}
>{\centering\arraybackslash}m{0.6cm}}  
\toprule
\textbf{Models} & \textbf{NHSR} & \textbf{\textit{Rel}} & \textbf{T$_{\text{LC}}$} &  \textbf{METEOR} & \textbf{BS.}\\
\hline
\multicolumn{6}{c}{\textit{\underline{Qwen2 Instruct 7B}}} \\
w/ SFT & 24.239\% & 5.179 & 8.592 & 0.317 & 0.712\\
+ HIPO M$_1$ & 24.441\% & 5.563 & 8.906 & 0.351 & 0.735\\
+ HIPO M$_2$ & 25.823\% & 5.844 & 8.880 & 0.359 & 0.741\\
+ HIPO M$_3$ & 27.435\% & 5.929 & 9.181 & 0.366 & 0.746\\
\hline
\multicolumn{6}{c}{\textit{\underline{GLM4 Chat 9B}}} \\
w/ SFT & 26.941\% & 5.832 & 8.771 & 0.329 & 0.744\\
+ HIPO M$_1$ & 32.406\% & 6.735 & 8.972 & 0.376 & 0.758\\
+ HIPO M$_2$ & 33.917\% & 6.889 & 8.975 & 0.396 & 0.761\\
+ HIPO M$_3$ & 38.353\% & 7.025 & 9.079 & 0.407 & 0.762\\
\bottomrule
\end{tabular}
\caption{Impact of HIPO iterations on LegalHalBench. M$_n$ represents the n-th iteration of HIPO. BS. is the abbreviation for BERTScore.}
\label{ResultsTable_HIPOIterations}
\end{table}

\paragraph{Comparison with RAG.}
We compare our method with retrieval augmentation generation (RAG), another technique to improve the factuality of LLMs. In the RAG experiments, we use BGE~\cite{bge_embedding} as the retriever, which uses the user's input as the query to retrieve the top three most probable legal statutes from an authentic legal corpus as additional knowledge for the LLM. As shown in Tab.~\ref{BGE}, introducing the BGE retriever yields performance improvement in terms of usefulness metrics, but the enhancement is limited. We speculate two possible reasons for this limited improvement: first, the BGE retriever may not accurately retrieve legal provisions highly relevant to the user's query; second, the external knowledge, namely the referenced legal statutes, that we provide to the LLM may conflict with its internal knowledge, resulting in suboptimal responses within the legal domain.

\begin{table}[t]
\centering
\small
\begin{tabular}{>{\raggedright\arraybackslash}m{3cm}>{\centering\arraybackslash}m{1.2cm}>{\centering\arraybackslash}m{0.7cm}>{\centering\arraybackslash}m{1cm}}  
\toprule
\textbf{Models} & \textbf{METEOR} & \textbf{BS.}  &\textbf{ROUGE} \\
\hline
\multicolumn{4}{c}{\textit{\underline{Open-source LLMs}}} \\
Llama3.1 70B & 0.200 & 0.707 & 0.221 \\
Llama3.1 70B w/ BGE & 0.236 & 0.719 & 0.254 \\
Llama3.1 405B & 0.232 & 0.726 & 0.249 \\
Llama3.1 405B w/ BGE & 0.219 & 0.720 & 0.260 \\
\hline
\multicolumn{4}{c}{\textit{\underline{Proprietary LLMs}}} \\
GPT4o-mini & 0.290 & 0.731 & 0.254 \\
GPT4o-mini w/ BGE & 0.334 & 0.734 & 0.267 \\
GPT4o & 0.348 & 0.739 & 0.266 \\
GPT4o w/ BGE & 0.337 & 0.739 & 0.281 \\
\hline
\multicolumn{4}{c}{\textit{\underline{Qwen2 Instruct 7B}}} \\
w/ BGE & 0.259 & 0.689 & 0.173 \\
w/ SFT + HIPO & \underline{0.366} & \underline{0.746} & \underline{0.303} \\
\hline
\multicolumn{4}{c}{\textit{\underline{GLM4 Chat 9B}}} \\
w/ BGE  & 0.295 & 0.717 & 0.182 \\
w/ SFT + HIPO & \textbf{0.407} & \textbf{0.762} & \textbf{0.340} \\
\bottomrule
\end{tabular}
\caption{Comparative results with RAG on different LLMs. Bold scores represent the best performance within the same model, while underlined scores represent the second best. BS. is the abbreviation for BERTScore.}
\label{BGE}
\end{table}


\begin{figure}[t]
    \centering
    \includegraphics[width=\columnwidth, trim=5 5 5 5, clip]{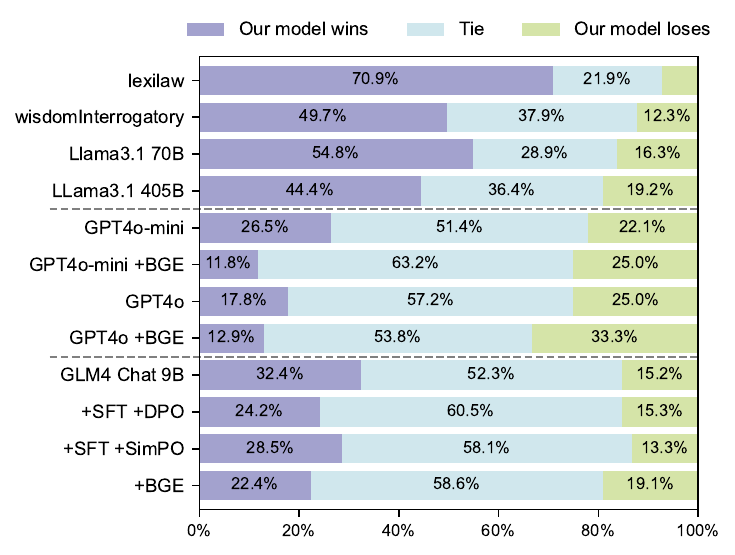}
    \caption{The win rate of the LegalHalBench experiment. The chart presents the win rates of GLM4 Chat 9B-based HIPO against other LLMs, evaluated using the latest GPT-4-turbo.}
    \label{fig:WinRate}
\end{figure}

\subsection{Paired Comparison Experiments on LegalHalBench}
As shown in Fig.~\ref{fig:WinRate}, our model significantly outperforms both existing specialized LLMs and open-source general-purpose LLMs, demonstrating a superior win rate. When compared to GPT4o, our model maintains an unbeaten rate of over 75\%. Furthermore, the performance of our HIPO strategy surpasses traditional DPO and SimPO in terms of win rates, underscoring the efficacy of the proposed approach.

To avoid position bias, the positions of the options under test are swapped, as GPT-4-turbo is used as the evaluator. If GPT-4-turbo does not maintain consistent preferences after the order of options is changed, these instances are recorded as draws. More details about win-rate computation can be found in Fig.~\ref{fig:WinRateTemplate}.

\subsection{Ablation Studies}
\label{sec:AblationStudies}

\begin{table}[t]
\centering
\small
\begin{tabular}{>{\centering\arraybackslash}m{1.5cm}>{\centering\arraybackslash}m{0.8cm}>{\centering\arraybackslash}m{0.6cm}>{\centering\arraybackslash}m{0.6cm}>{\centering\arraybackslash}m{1cm}>{\centering\arraybackslash}m{0.6cm}}
\toprule
\textbf{Methods} & \textbf{NHSR} & \textbf{\textit{Rel}} & \textbf{T$_{\text{LC}}$} & \textbf{METEOR} & \textbf{BS.} \\
\midrule
Ours & 38.353\% & 7.025 & 9.079 & 0.407 & 0.762 \\
\\[-2ex]\hline\\[-2ex]
w/ DPO loss & 32.877\% & 6.705 & 9.032 & 0.375 & 0.749 \\
\\[-2ex]\hline\\[-2ex]
w/o Hard Sample & 28.501\% & 6.535 & 8.837 & 0.382 & 0.750 \\
\\[-2ex]\hline\\[-2ex]
w/o Iteration & 26.941\% & 5.832 & 8.771 & 0.329 & 0.744 \\
\bottomrule
\end{tabular}
\caption{Ablation studies. BS. is the abbreviation for BERTScore.}
\label{AblationTable}
\end{table}

As shown in Tab.~\ref{AblationTable}, we conduct ablation studies to evaluate the contribution of each component in our proposed method. The first row presents the performance of our full method. In the second row, we replace our modified loss function with the original DPO loss function while keeping the rest of the HIPO strategy intact. This results in decreased performance across all metrics, indicating the effectiveness of our modified loss function. In the third row, omitting hard sample-aware selection during training results in a further performance decline, with NHSR decreasing to 28.501\%. In the last row, removing iterative DPO leads to significant declines across all performance metrics.

\subsection{Human Consistency Experiments on Legal Hallucination Evaluation Metrics}
\label{sec:HumanConsistencyExperimentsonLegalHallucinationEvaluationMetrics}

Although some experimental results~\cite{zheng2023judging} indicate that powerful LLMs like GPT-4 achieve over 80\% consistency with human judgments, effectively evaluating the performance of generative models still requires assessing their alignment with human-annotated data.

We use accuracy to measure the alignment between the Non-Hallucinated Statute Rate and human judgments. Additionally, we utilize Spearman's rank correlation coefficient and Pearson correlation coefficient to assess the consistency between the LLM-generated Statute Relevance scores and human ratings, as well as the consistency between the LLM-generated Legal Claim Truthfulness scores and human ratings.

For the Non-Hallucinated Statute Rate metric, we randomly sample 100 QA pairs from the inference results of various models. These 100 samples are then reviewed by three independent lawyers, who are asked to assign a binary label to each sample, indicating whether the statute referenced in the sample is hallucinated. The majority label among the three lawyers is used as the golden answer. We then calculate the accuracy to assess the alignment between the Non-Hallucinated Statute Rate and the golden answers. We interpret a higher accuracy as an indication of greater consistency with human preferences.

In assessing the consistency of the Statute Relevance Rate and Legal Claim Truthfulness metrics, we randomly select 100 QA pairs, which are independently reviewed by three lawyers. Each lawyer assigns scores to the samples based on specific evaluation criteria, ensuring the fairness of the assessment process through independent scoring. The final human score is obtained by averaging the scores from all three lawyers.

As shown in Tab.~\ref{HumanConsistencyExperiment}, the Non-Hallucinated Statute Rate, Statute Relevance Rate, and Legal Claim Truthfulness metrics demonstrate strong alignment with human judgment in evaluating hallucinations in LLMs within the legal domain.

\begin{table}[t]
    \centering
    \small
    \setlength{\tabcolsep}{4pt}
    \begin{tabular}{ccccc}
        \toprule
        \textbf{NHSR} & \multicolumn{2}{c}{\textbf{\textit{Rel}}} & \multicolumn{2}{c}{\textbf{T$_{\text{LC}}$}} \\
        \cmidrule(lr){1-1} \cmidrule(lr){2-3} \cmidrule(lr){4-5}
        \textbf{Acc} & \textbf{$\rho$} & \textbf{PCCs} & \textbf{$\rho$} & \textbf{PCCs} \\
        \midrule
        98\% & 0.820 & 0.821 & 0.617 & 0.707\\
        \bottomrule
    \end{tabular}
    \caption{Human Consistency Experiment.$\rho$ refers to Spearman's rank correlation coefficient, and PCCs is the abbreviation for Pearson correlation coefficients.}
    \label{HumanConsistencyExperiment}
\end{table}

\section{Conclusion}
In this paper, we first introduce LegalHalBench, a benchmark designed to assess hallucinations in legal question answering applications of LLMs. This benchmark enables precise evaluation through tailored metrics for different hallucination types. We then propose a novel two-phase training model that integrates SFT with HIPO to enhance the factual accuracy and relevance of LLM responses. Extensive experiments validate the effectiveness of our training method.

\section*{Limitations}
Although our research indicates that using the prposed HIPO can make the responses of baseline model less hallucinatory and more helpful, we observe that as the number of iterations increases, the model's performance growth slows down, and eventually the performance tends to plateau. Continuing to increase the number of iterations may not only fail to enhance the model's performance but could also degrade its capabilities in certain areas. Our future research focuses on exploring whether more rounds of iterations can continuously improve the model's performance.

Additionally, we use the presence of hallucinations in the model's responses to previous training data as a criterion for judging if the model has learned the knowledge. In reality, defining whether a model has truly learned the knowledge is a complex issue worth further investigation. Another important area we plan to explore in the future is to define the knowledge boundaries of LLMs.

\section*{Ethics Statement}
Given the sensitive nature of the legal domain, the application of artificial intelligence technology in this field requires careful management. To address ethical concerns, we undertake the following initiatives. First, to prevent the leakage of private information such as real names by the model, we anonymize or replace sensitive information (such as personal names) with third-person references when constructing the training dataset and benchmark. Second, to prevent the model from generating biased outputs, we mask parts of discriminatory data with "**".

\section*{Acknowledgements}
This work was supported by the National Natural Science Foundation of China (62441605, 62376243), and the Starry Night Science Fund at Shanghai Institute for Advanced Study (Zhejiang University). We would like to thank the anonymous reviewers for their comments and suggestions.

\bibliography{COLING2025}
\bibliographystyle{acl_natbib}

\appendix

\section{Appendix}
\label{sec:appendix}

\subsection{Exploration of Training Data Quality}
\label{sec:ExplorationofTrainingDataQuality}
\subsubsection{Capability of Correcting Fabricated Statutes}
\label{sec:CapabilityofCorrectingFabricatedStatutes}
To verify whether replacing fabricated statutes with real statutes that are semantically most similar can maintain a high level of accuracy in our training data construction method (Sec.~\ref{sec:QAforpubliclydatasets}), we design the following experiment.

\textbf{(1) Problem Collection and Model Inference:} We randomly sample 100 challenging legal questions from the real world, ensuring that these questions cover both civil law and criminal law. Additionally, we ensure that the statutes referenced in these questions cannot be fully and correctly inferred by existing LLMs. The 100 questions are then provided to GPT-4-turbo, Llama 3.1-405B, and GLM4 Chat 9B, and the models generate their corresponding original answers.

\textbf{(2) Semantic Search and Statute Replacement:} We extract the statutes mentioned in the model-generated responses (\(L_{\text{resp}}\)) and conduct a semantic search in our local database to identify the most semantically similar statutes (\(L_{\text{search}}\)). We then replace the references to \(L_{\text{resp}}\) in the original answers with \(L_{\text{search}}\), producing the revised answers.

\textbf{(3) Manual Verification:} We hire three legal practitioners to independently verify whether \(L_{\text{search}}\) correctly replaces \(L_{\text{resp}}\). Specifically, each lawyer is provided with 300 triplets of \{question, original answer, revised answer\} and asked to manually label each triplet with a binary classification to assess whether the statute replacement is accurate.

\begin{table}
\centering
\small
\begin{tabular}{>{\centering\arraybackslash}m{2cm}>{\centering\arraybackslash}m{3cm}}  
\toprule
\textbf{Models} & \textbf{Error correction success rate}\\
\midrule
GPT4o & 84\% \\
Llama3.1 405B & 74\% \\
GLM4 Chat 9B & 83\% \\
\bottomrule
\end{tabular}
\caption{Error correction success rate.}
\label{Errorcorrectionsuccessrate}
\end{table}

According to Tab.~\ref{Errorcorrectionsuccessrate}, this method effectively corrects the statutes generated by the models, achieving an average error correction success rate of over 80\%.

\subsubsection{Manual Evaluation of Training Data Quality}
\label{sec:ManualEvaluationofTrainingDataQuality}

To provide a more intuitive assessment of the quality of the training data we generate, we design the following experiment.

We randomly sample 100 cases from the legal QA dataset and hire three lawyers to manually assess these 100 samples. Specifically, the evaluation is conducted using a ``veto'' system. There are three evaluation criteria, and if any one (or more) of these criteria is not met, the sample is labeled as low-quality. Conversely, samples that meet all three criteria are considered high-quality. The three criteria are: correct legal provisions, helpful legal advice, and practical applicability of the training data.

According to the results of the manual evaluation, less than 4\% of the data is labeled as low-quality. The reason why the proportion of low-quality data is much smaller than the error rate in the correction process is that Sec.~\ref{sec:CapabilityofCorrectingFabricatedStatutes} represents an idealized extreme scenario, where all legal provisions generated by the LLM need correction, and \(L_{\text{search}}\) directly replaces \(L_{\text{resp}}\). In reality, LLMs do possess some ability to generate accurate legal provisions. Moreover, during data construction, before replacing \(L_{\text{resp}}\) with \(L_{\text{search}}\), we also use the LLM to verify whether \(L_{\text{resp}}\) is relevant to the query. This additional step further enhances the reliability of our dataset.

In summary, our proposed approach can effectively replace most fabricated legal provisions with accurate ones. Therefore, we believe that the advantages of this automated, low-cost method for constructing high-quality training data outweigh its potential shortcomings.

\subsection{Extracting Statutes Using LLMs}
\label{sec:ExtractingStatutesUsingLLMs}

\begin{figure}[ht]
    \centering
    \includegraphics[width=\columnwidth, trim=10 10 10 10, clip]{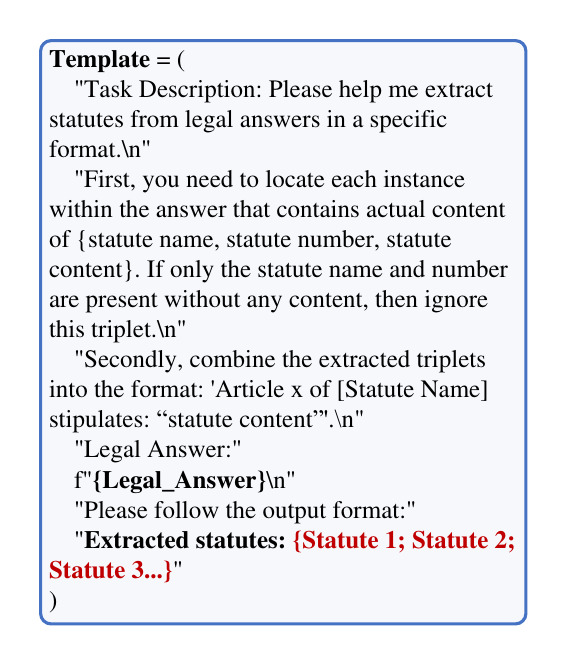}
    \caption{Template for Extracting Statutes Using LLMs.}
    \label{fig:TemplateForExtracting}
\end{figure}

Not all LLMs exhibit robust instruction-following capabilities; as a result, not every LLM can generate legal provisions in the specified format. When legal provisions cannot be directly extracted using regular expressions, we use LLMs (such as GPT-4-turbo) to assist in extracting and formatting the information as required. Refer to Fig.~\ref{fig:TemplateForExtracting} for the template used in this extraction process.

\subsection{Technical Details of NHSR}
\label{sec:TechnicalDetailsofNHSR}
This metric can be calculated as follows:

(1) Extract the generated statutes using regular expressions or LLMs, denoted as \( L_{\text{gen}} = \{S_{\text{name}}, S_{\text{number}}, S_{\text{content}}\} \). For detailed information on using LLMs to extract statutes, please refer to Sec.~\ref{sec:ExtractingStatutesUsingLLMs}.

(2) Embed the extracted content using the SimBERT~\cite{SimbertBaseChinese} model alongside all contents from a real statute database.

(3) Calculate the semantic similarity between the generated content and the real statutes, selecting the real statute with the highest similarity as the response. This returned statute is denoted as \( L_{\text{best}} = \{S'_{\text{name}}, S'_{\text{number}}, S'_{\text{content}}\} \). The rationale behind this process is that although large models do not have the capability of generating completely hallucination-free statutes, they typically maintain high semantic consistency with real statutes. The closer the model-generated statute is to a real statute, the higher its semantic similarity.

(4) Using rule-based comparisons, assess whether \( L_{\text{gen}} \) and \( L_{\text{best}} \) indicate hallucinations in the model-generated statutes. Specifically, we consider the generated statute \(L_{\text{gen}}\) to be non-hallucinated if \(S'_{\text{content}}\) is a subset of \(S_{\text{content}}\), \(S'_{\text{number}}\) equals \(S_{\text{number}}\), and \(S'_{\text{name}}\) falls within a specific set of appellations for \(S_{\text{name}}\).

\subsection{Technical Details of Statute Relevance Rate}
\label{sec:TechnicalDetailsofStatuteRelevanceRate}

\begin{figure}[ht]
    \centering
    \includegraphics[width=\columnwidth, trim=10 10 10 10, clip]{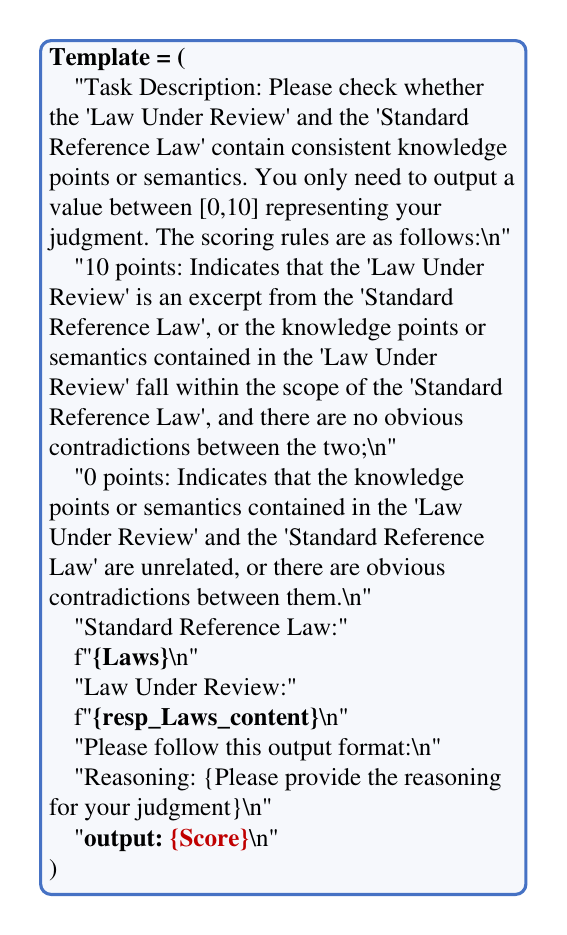}
    \caption{Prompt for the Statute Relevance Rate.}
    \label{fig:rel1}
\end{figure}

\begin{figure}[ht]
    \centering
    \includegraphics[width=\columnwidth, trim=10 10 10 10, clip]{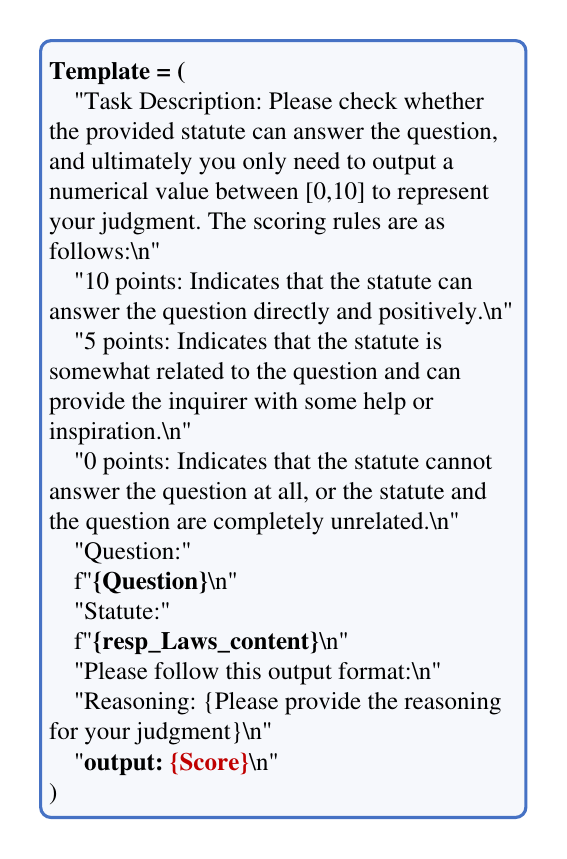}
    \caption{Prompt for the Statute Relevance Rate.}
    \label{fig:rel2}
\end{figure}

NHSR is not the preferred metric in the following two scenarios: (1) The generated statute contains no hallucination but is not highly relevant to the user's question; (2) The generated statute includes some inaccuracies yet still provides correct legal knowledge that can be helpful to the user. Therefore, we need to propose a metric to evaluate the relevance of the knowledge contained in the generated statutes.

We develop two distinct sets of prompts for evaluating statutes generated by the LLM, tailored to whether the statutes contain hallucinations. Fig.~\ref{fig:rel1} illustrates the prompts used when the statutes are free from hallucinations. In this scenario, we directly employ the LLM to assess the relevance between the generated statute and the user's query, thereby evaluating the utility of the knowledge contained within the statute. Fig.~\ref{fig:rel2} displays the prompts utilized when the statutes include hallucinations. In this case, we use the LLM to examine the consistency of knowledge points between the generated statute and a reference statute, to determine the relevance of the generated content.

The metric is computed as follows:
\begin{equation}
    Rel = \frac{\sum_{i=1}^N s_i}{N}
\end{equation}

\subsection{Technical Details of Legal Claim Truthfulness}
\label{sec:TechnicalDetailsofLegalClaimTruthfulness}
\begin{figure}[ht]
    \centering
    \includegraphics[width=\columnwidth, trim=10 10 10 10, clip]{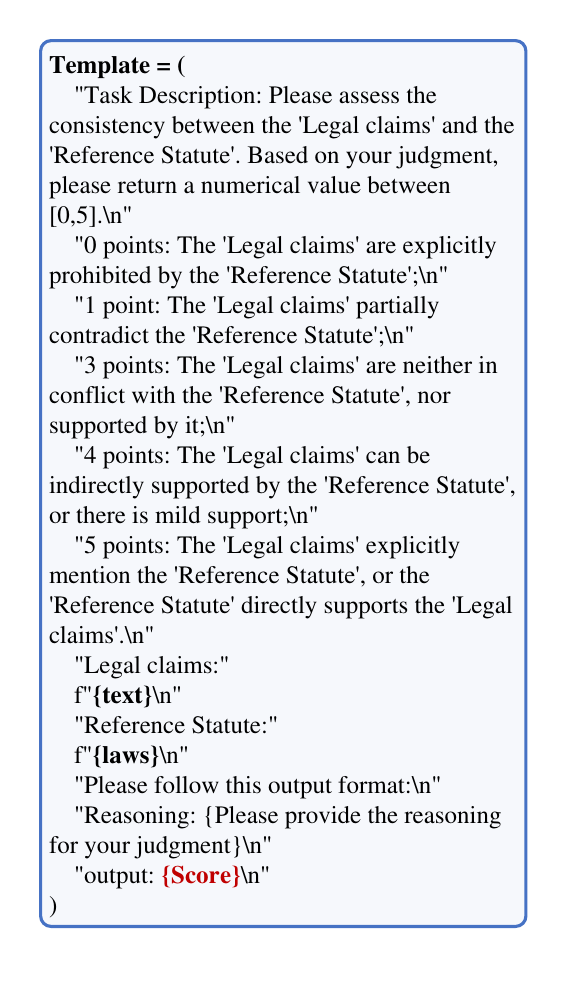}
    \caption{LLM prompts for Legal Claim Truthfulness.}
    \label{fig:yizhixing}
\end{figure}

We extract the suggestion component from the responses generated by the LLM and then employ the LLM to assess the legality of these suggestions. GPT-4-turbo is asked to return a scalar value \( s_i \), where \( s_i \) ranges from 0 to 5. For convenience in subsequent statistics, we directly double the value of \( s_i \), so the resulting values fall within the range [0,10]. The prompts used for the LLM are illustrated in Fig.~\ref{fig:yizhixing}.

The calculation formula is as follows:
\begin{equation}
    T_{\text{LC}} = \frac{\sum_{i=1}^N s_i}{N}
\end{equation}

\subsection{Win-rate Template}
\label{sec:A5}

\begin{figure*}[b]
    \centering
    \includegraphics[width=\textwidth, trim=10 10 10 10, clip]{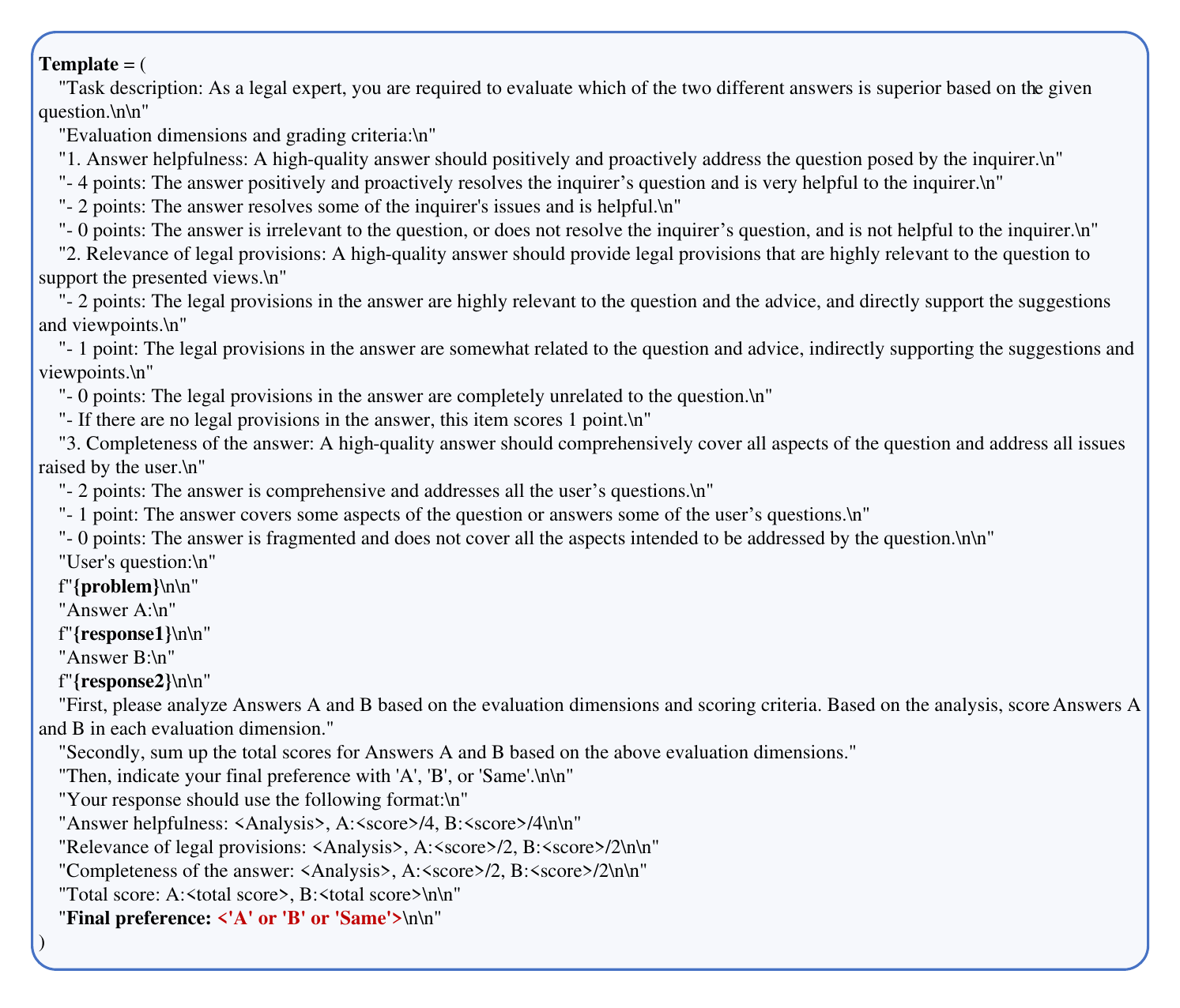}
    \caption{Win-rate Template}
    \label{fig:WinRateTemplate}
\end{figure*}

In the evaluation guide, we instruct GPT-4-turbo to score the win-rate based on three dimensions (Fig.\ref{fig:WinRateTemplate}), and we ensure that the temperature of GPT-4-turbo is set to 0:

\begin{itemize}
    \item \textbf{Helpfulness of the Answer (4 points):}
    We believe that high-quality legal responses should positively and proactively address the question raised by the inquirer. We consider this dimension the most important, so it is given the full 4 points. The specific criteria are as follows: (1) The answer positively and proactively solves the inquirer's question, being very helpful to the inquirer, scoring 4 points. (2) The answer partially solves the inquirer's question, being somewhat helpful to the inquirer, scoring 2 points. (3) The answer is unrelated to the question or does not solve the inquirer's question, being not helpful to the inquirer, scoring 0 points.
    
    \item \textbf{Relevance to Legal Regulations (2 points):}
    High-quality legal answers should provide legal regulations that are highly relevant to the question to support the points made. In this dimension, we do not consider whether the law is real or hallucinated; we only look at its relevance to the question and the answer. The specific criteria are as follows: (1) The legal provisions in the answer are highly relevant to the question and suggestions, directly and strongly supporting the suggestions and viewpoints, scoring 2 points. (2) The legal provisions in the answer are somewhat relevant to the question and suggestions, indirectly supporting the suggestions and viewpoints, scoring 1 point. (3) The legal provisions in the answer are completely unrelated to the question, scoring 0 points. If the answer does not contain any legal provisions, it scores 1 point.
    
    \item \textbf{Completeness of the Answer (2 points):}
    High-quality legal responses should comprehensively cover all aspects of the question and address all issues raised by the user. The specific criteria are as follows: (1) The answer is comprehensive and addresses all of the user's questions, scoring 2 points. (2) The answer covers some aspects of the question or addresses some of the user's questions, scoring 1 point. (3) The answer is scattered and does not cover all the aspects that the question aims to solve, scoring 0 points.

\end{itemize}

We require GPT-4-turbo to provide its evaluation in a step-by-step reasoning format. According to our experiments, step-by-step reasoning is more objective and accurate than directly giving scores. The specific reasoning process is as follows:

(1) We require GPT-4-turbo to analyze each dimension of the two candidate answers.

(2) Based on the analysis, GPT-4-turbo provides the scores.

(3) The scores for each evaluation dimension are summarized to obtain the total scores for Answer A and Answer B.

(4) Finally, based on the total scores of the candidate answers, the preference is determined.

\subsection{Generating Legal Questions Using GPT-4-turbo}
\label{sec:GeneratingLegalQuestionsUsingGPT4turbo}

In Sec.~\ref{sec:QuestionConstruction}, to enable the model to learn about a specific legal provision \(L\), we need to construct a set of questions related to the specific legal provision \(L\).

\textbf{(1) Provide the legal provision and set the task:} Provide the legal provision \( L \) to GPT-4-turbo and instruct it to assume the role of a law professor, designing multiple questions that can be answered based on this legal provision. For this task, we set GPT-4-turbo's temperature parameter to 0.7 to ensure that the generated questions are both creative and reasonable.

\textbf{(2) Impose constraints on the questions:} Impose constraints on the questions to ensure they are reasonable and answerable within the framework of the provided legal provision. Each question should be contextualized rather than being a simple query. Additionally, the phrasing should align with the vocabulary and style commonly used in modern Chinese.

\textbf{(3) Design diverse question formats:} Design a range of question formats to enhance the diversity of the generated questions. Examples include using a ``first-person perspective with subjective emotion,'' a ``first-person perspective maintaining objective neutrality,'' and a ``colloquial style,'' among others, to create a variety of question forms and styles.

\subsection{Generating High-Quality Responses Based on Given Legal Materials}
\label{sec:GeneratingHigh-QualityResponsesBasedonGivenLegalMaterials}

In Sec.~\ref{sec:QuestionConstruction}, we provide GPT-4-turbo with a legal question \( Q \) and the relevant legal provision \( L \), asking it to generate a response. The process is as follows: First, we provide the reference legal provision \( L \) to GPT-4-turbo and instruct it to act as a legal expert, using the provision to answer the question. We set GPT-4-turbo's temperature to 0.7. 

To ensure the response's quality, we impose specific constraints on GPT-4-turbo's output through prompt guidelines. These guidelines require GPT-4-turbo to first summarize and clarify the question, identifying the key points that need addressing. It should then use the provided legal provision \( L \) to answer the question. If the provision does not fully address the issue, GPT-4-turbo should supplement the answer with additional relevant information to improve completeness and usefulness.

\subsection{Detailed Introduction to Baseline}
\label{sec:DetailedIntroductiontoBaseline}

\begin{itemize}
    \item \textbf{Llama 3.1-70B:} Llama 3.1 70B shows comprehensive improvements over its predecessor, natively supporting 8 languages with a maximum context window of 128k.
    
    \item \textbf{Llama 3.1-405B:} Llama 3.1 405B is trained on 150 trillion tokens (equivalent to 750 billion words), with fine-tuning on 25 million synthetic data. Llama 3.1 405B is fully competitive with the most advanced proprietary models in various tasks.
    
    \item \textbf{GPT4o-mini:} GPT-4o mini is the mini version of OpenAI’s GPT-4o, featuring low cost and rapid response capabilities, suitable for a variety of application scenarios.
    
    \item \textbf{GPT4o:} GPT-4o is the latest commercial model developed by OpenAI, with performance in English text and code comparable to GPT-4-turbo.
    
    \item \textbf{wisdomInterrogatory:} Based on the Baichuan 7B architecture, this model undergoes secondary pre-training using 40GB of legal data. During the instruction fine-tuning phase, 100k of instruction fine-tuning training data is used.
    
    \item \textbf{LexiLaw:} This model, based on the ChatGLM-6B architecture, is fine-tuned with a variety of legal and general domain datasets. It includes BELLE 1.5M general domain data, 144k legal QA from LawGPT\_zh and Baidu Zhidao, law exam and directive data from Lawyer LLaMA, and 20k high-quality QA data from Hualaw, supplemented with laws, regulations, and legal reference texts.
    
    \item \textbf{Qwen2-Instruct-7B:}  Qwen2-7B-Instruct supports context lengths of up to 131,072 tokens, based on the Transformer architecture.
    
    \item \textbf{GLM4-Chat-9B:} GLM4-9B is the open-source version in the latest generation of the GLM-4 series of pre-trained models.
\end{itemize}

The retrieval system we use is the BAAI General Embedding (BGE), specifically version bge-base-zh-v1.5. BGE is suitable for tasks such as text similarity, ranking, question answering, and retrieval across multiple languages.

\subsection{Evaluating Model Performance on LawBench}
\label{sec:EvaluatingModelPerformanceonLawBench}
In Tab.~\ref{Lawbench}, our model demonstrates superior performance in the tasks of Case Analysis and Legal Consultation on the Lawbench dataset, outperforming its baseline model, GLM4-Chat-9B, and surpassing all other open-source models. The improvement in accuracy for single-choice questions can be attributed to the model’s effective acquisition of a substantial amount of {scenario, statute} matching information pairs. This information plays a key role in analyzing and accurately answering single-choice questions, which contributes to the observed performance increase.

\begin{table}[htbp]
\centering
\small
\setlength{\tabcolsep}{1pt}
\begin{tabular}{m{3cm}<{\centering}m{2cm}<{\centering}m{2cm}<{\centering}}  
    \toprule
    \textbf{Methods} & \textbf{Case analysis} (Acc) & \textbf{Legal consultation} (ROUGE-L)  \\
    \midrule
    lawgpt-7b-beta1.1-hf &9.20 &7.62  \\
    lexilaw-6b-hf &28.60 &15.82  \\
    lawyer-llama-13b-hf &26.60 &16.94  \\
    fuzi-mingcha-7b-hf &20.00 &16.64  \\
    chatlaw-13b-hf &28.80 &17.17  \\
    chatlaw-33b-hf &34.20 &16.55   \\
    wisdomInterrogatory &25.40 &18.26   \\
    GPT-3.5-turbo-0613 &27.40 &17.45   \\
    GPT4 &48.60 &19.65   \\
    GLM4-Chat-9B &54.40 &15.11   \\
    Ours M$_3$ &63.00 &18.87   \\
    \bottomrule
\end{tabular}
\caption{Results on Lawbench.} 
\label{Lawbench}
\end{table}

\subsection{Statistical Distribution of LegalHalBench}
\label{sec:StatisticalDistributionofLegalHalBench}

\begin{table}[htbp]
\centering
\small
\begin{tabular}{>{\raggedright\arraybackslash}m{1.9cm} 
                >{\centering\arraybackslash}m{1.2cm} 
                >{\centering\arraybackslash}m{1.2cm} 
                >{\centering\arraybackslash}m{1cm}}  
\toprule
 & \textbf{Civil Law} & \textbf{Criminal Law} & \textbf{Total} \\
\midrule
Legal Provisions & 750 & 85 & 835 \\
\midrule
Questions & 1,384 & 604 & 1,988 \\
\midrule
Avg. Length of Question & 172.52 & 299.37 & 211.06 \\
\midrule
Avg. Length of Answer & 471.43 & 295.20 & 417.89 \\
\bottomrule
\end{tabular}
\caption{Statistical Distribution of LegalHalBench.}
\label{DataDistributionofLegalHalBench}
\end{table}

Tab.~\ref{DataDistributionofLegalHalBench} shows the Statistical Distribution of LegalHalBench.

\subsection{Implementation Details}
\label{sec:ImplementationDetails}

In the first phase of training, based on Qwen2-Instruct-7B and GLM4-Chat-9B, we conduct two rounds of SFT with a learning rate of 5e-6. In the second phase, we perform three cycles of HIPO while maintaining the same initial learning rate of 5e-6. We use both SFT and HIPO under the LoRA framework~\cite{hu2021lora}. With a single A100 40G GPU, training one epoch of SFT takes approximately 1 hour, while training one full epoch of HIPO takes around 3.5 hours. For the traditional DPO, we adopt the same parameter settings as HIPO. For SimPO, we utilize the parameter settings described in the paper~\cite{meng2024simpo}.

\subsection{Cost of Constructing the LegalHalBench}
\label{sec:CostofConstructingtheLegalHalBench}

We provide the needed cost details as follows: Initially, we use the GPT4-turbo API to answer 1,988 questions, costing approximately \$111. In the second phase, four lawyers review and refine the responses, followed by a final review by two senior lawyers to ensure accuracy. Overall, this process requires over 200 hours of professional lawyer time.

\end{document}